\documentclass[preprint,authoryear,12pt]{elsarticle}

\usepackage{graphicx}
\usepackage[english]{babel}

\usepackage{amssymb}

\journal{Pattern Recognition Letters}

\begin{document}

\begin{frontmatter}



\title{Image denoising assessment using anisotropic stack filtering}


\author{Salvador Gabarda and Gabriel Crist\'obal}

\address{Instituto de Optica, Serrano 121, 28006 Madrid, Spain}

\begin{abstract}
In this paper we propose a measure of anisotropy as a quality parameter to estimate the amount of noise in noisy images. The anisotropy of an image can be determined through a directional measure, using an appropriate statistical distribution of the information contained in the image. This new measure is achieved through a stack filtering paradigm. First, we define a local directional entropy, based on the distribution of 0's and 1's in the neigborhood of every pixel location of each stack level. Then the entropy variation of this directional entropy is used to define an anisotropic measure. The empirical results have shown that this measure can be regarded as an excellent image noise indicator, which is particularly relevant for quality assessment of denoising algorithms. The method has been evaluated with artificial and real-world degraded images.

\end{abstract}


\end{frontmatter}


\section{Introduction}
\label{intro}
Image anisotropy with respect to a given image feature is the property of having a directional dependence. This property is conceptually the opposite to isotropy that implies directional uniformity. In a previous paper \citep{Gabarda07}, we have shown that the entropy calculated in a directional way can be used to define a measure of anisotropy that maintains a strong correlation with the image noise content. In this paper, a new measure of anisotropy is defined for the binary signals generated using the stack filtering paradigm. The purpose of such technique is for assessing the quality of noise reduction methods. Many image denoising methods have been proposed in the literature. Here we focus this study on a particular type of non-linear  filters called stack filters. Stack filters (SFs) \citep{Astola_97,Coyle_89,Coyle_88} have been proposed as a useful approach for noise reduction with many interesting properties particularly in the case of correlated noise. 
Stack filters have been used with success for different filtering purposes. Especially interesting is the contribution due to  \citep{Coyle_88} for introducing a filtering paradigm which minimizes the mean absolute error between its output and a desired signal, given noisy observations of the signal as the filter's input. These results show that optimal stack filtering under the mean absolute error criterion is analogous to optimal linear filtering under the mean squared error criterion. In \citep{Coyle_89}, structural and estimation approaches for images are combined into a methodology that provides the best filter from a very large class of generalized stack filters. In \citep{Buemi_10}, the behavior of adaptive stack filters on synthetic aperture radar (SAR) was evaluated. A classification of simulated and real degraded SAR images is carried out using a stack filtering trained with selected samples. The results of a maximum likelihood classification of these data are evaluated and compared with the results using the Lee and the Frost classical filtering approach.
In this paper, our aim is focused on the use of SFs for quality evaluation rather than defining a new image denosing method. We define a new measure of anisotropy for the binary signals generated by the stack filters paradigm. This measure is used here to define a technique for assessing the quality of noise reduction methods in general.
This paper is organized as follows, Section 2 gives the mathematical background required for understanding the proposed measure, Section 3 presents some application examples and finally conclusions are drawn in Section 4.

\section{Mathematical background}
\label{back}
\subsection{Stack filtering}
A recent description of stack filters applied to image processing may be found in \citep{Buemi_10}. SFs present two main properties. The first one is threshold decomposition (TD) and the other one is known as the stacking property \citep{Coyle_88}. To underline the threshold decomposition property let us suppose that $X=(x_{1},...,x_{n})$ is a vector representing a gray-scale image where each element $x_{i}$ corresponds  to one pixel. We can define a thresholding function $T$ and consider that $T_{i}(X)$ is the thresholded binary vector determined by this function for a threshold level $l$. Analogously, $T_{l}(x_{i})$ is the thresholded binary value of element $x_{i}$, defined as $T_{l}(x_{i})=1 \hspace*{3mm} if \hspace*{3mm} x_{i} \geq l $ and $T_{l}(x_{i})=0 \hspace*{3mm} if  \hspace*{3mm} x_{i} < l$. This relationship implies that the gray-scale image $X$ can be recovered from the set of binary thresholded images by addition, that is, $x_{i}=\sum_{l=1}^{L}T_{l}(x_{i})$, where $L$ is the number of possible levels in the gray-scale image. The main advantage of a filter $F$ having the threshold decomposition property is based on considering that the output of such a filter $F(X)$ is the result of three operations, namely a decomposition into thresholded binary images, secondly it comes a filtering process in this binary state and finally an addition of the binary products is performed. The total filtering process can be summarized as $F(X)=\sum_{l=1}^{L}F(T_{l}(X))$. Here $F(T_{l}(X))$ represents the filtering operation over the binary image $T_{l}(X)$. This threshold decomposition opens up the possibility of realizing Boolean operations over the binary products of the image. The stacking property assumes that given a filter $F$ and two multilevel input vectors $U=(u_{1},...,u_{n})$ and $V=(v_{1},...,v_{n})$, the filter $F$ possesses the stacking property if $S(U) \geq S(V) \Leftrightarrow U \geq V$ holds. SFs have the stacking property, provided that reconstruction of the gray-scale image after the binary filtering process is achieved by adding 0 or 1 for each pixel position and level. 

For the current purpose, we present here a practical procedure that can be applied to 8-bit images. Let us suppose that image X is represented by a  $N{\times}M$ matrix, where each pixel gray-value is represented by $x_{k}$ , which is an integer number in the range from 0 to 255. In such circumstances, we can define a stack of  $L=255$  binary signals  $B^{l} (L=1,...,L)$, whose elements $b_{k}^{l}$ have a value given by
\begin{equation}
b_{k}^{l}=\left\{\begin{array}{rl}
	1  & \mbox{  if  $x_{k}\geq l$} \\
	0 & \mbox{ otherwise}
	   \end{array} \right.	  	
\end{equation}
and $k=1,...,M{\times}N$ indicate the pixel positions in the image. 

\subsection{Anisotropy stack level}
Without lack of generality let us suppose that we identify each of the  $B^{l}$ signals by $B$  and now we define a $d{\times}d$ ($d$ being  an odd number) squared matrix $D_{\theta}$ , whose elements are all zero except those who have the minimum Euclidean distance to the theoretical line with direction $\theta$ in Cartesian coordinates, referred to the central element of matrix $D_{\theta}$  (see Fig. 1 for an example).

\begin{figure}[h!]
  \centering
      \includegraphics[width=0.8\textwidth]{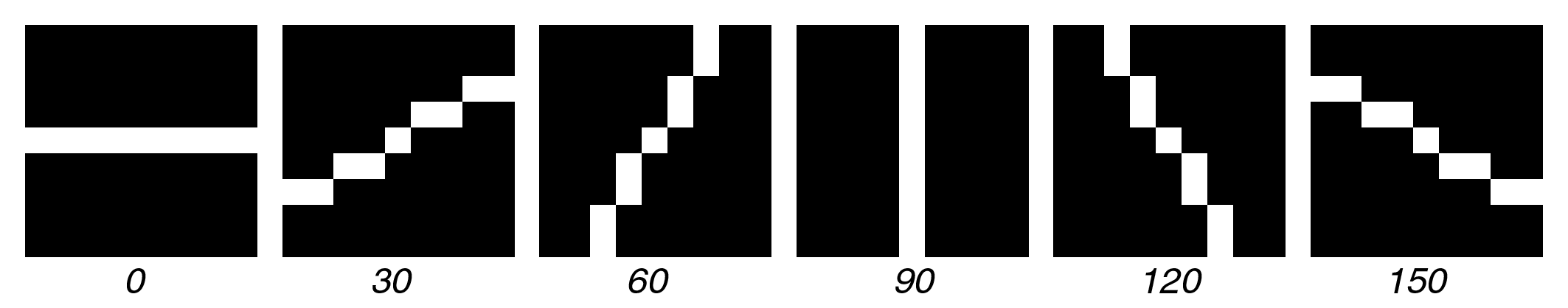}
  \caption{Different configuration of pixels to define directional filters $D_{\theta}$ , with  $N = 9$ pixels and six equally spaced orientations in degrees from 0 to 150 ( note that periodicity is  $\pi$, provided that 180 overlays 0.}
\end{figure}

After the previous assumptions, let us consider the following expression,

\begin{equation}
P_{l}^{\theta}=\frac{1}{d}(B*D_{\theta})
\end{equation}

It gives a matrix $P_{l}^{\theta}$ whose elements represent the probability of 1's in the oriented neighborhood of  each homologue pixel in matrix $B$. Here the symbol $*$  indicates the convolution operation.
The interpretation of Eq. 2 is straightforward by considering the convolution of a binary image $B$ with an spatial filter $D$. By definition,  $P(x,y)=\frac{1}{d}B*D=\frac{1}{d}\sum_{s=-a}^{a}\sum_{t=-b}^{b}D(s,t)B(x+s,y+t)$. Each pixel in $B$ is replaced by a weighted average of 1's in its neighborhood. In Eq. (2), this number represents the average number of ones in the diagonal, provided that outside the diagonal all values are zero. Hence, $P(x,y)$  stores a number equal to the rate of 1's in the diagonal neighborhood of  $B(x,y)$, equal to the probability of 1's in such region.
Consequently, $P_{0}^{\theta}=\bold{1}-P_{l}^{\theta}$ is the probability of 0's in the homologue neighborhood and $\bold{1}$ is a $N{\times}M$ matrix where all its elements are equal to 1.
Let call $p_{l}^{k,\theta}=p_{l}^{\theta}(b_{k})$ and $p_{0}^{k,\theta}=p_{0}^{\theta}(b_{k})$ the elements of $P_{l}^{\theta}$ and $P_{0}^{\theta}$ respectively, a local directional measure of entropy can be applied to the binary signals as follows:

\begin{equation}
r^{k,\theta}=\frac{1}{1-\alpha}\log_{2}\sum_{i=0}^{i=l} (p_{i}^{k,\theta})^{\alpha}
\end{equation}

This equation follows after applying to our distribution the generalized R\'enyi entropy \citep{Renyi76}. It is worthy to be noted that the R\'enyi entropy reverts to Shannon entropy \citep{Shannon_49} when $\alpha \rightarrow 1$. Our preference to use the R\'enyi entopy instead of the Shannon entropy is based primarily on its frequent use by researchers on space-frequency image analysis based on this approximation. Also the existence of a free parameter $\alpha$ gives an interesting flexibility for measuring entropic values. When $\alpha \rightarrow \infty$, this entropy seems to consider only events with the highest probability. Oppositely, small values of $\alpha$ tend to consider events more equally, regardless of their probabilities. Hence, we prefer to use the R\'enyi entopy rather than the Shannon entropy for image processing, due to its generalized character and the possibility of using $\alpha$ as a shape controlling parameter.
This pixel-wise directional entropy can be constrained to a scalar value for the whole binary level $B$ by averaging all $r^{k,\theta}$  values. Hence, 

\begin{equation}
\bar r^{\theta}=\langle r^{k,\theta} \rangle= \sum_{k}r^{k,\theta}
\end{equation}

Now, the anisotropy $A$ of a binary signal $B$ can be estimated by the variation in the set of directional entropies calculated by Eq. 4. We have considered here that a suitable statistical parameter to be associated with the image anisotropy is the standard deviation of the image directional entropy, providing that this parameter correlates well with the amount of variation of entropy when measured in different orientations \citep{Gabarda07}. Namely

\begin{equation}
A=\sigma=\sqrt{\frac{1}{T}\sum_{\theta=\theta_{l}}^{\theta=\theta_{T}}(\bar r^{\theta}-\mu)^2}
\end{equation}

where $\mu=\langle \bar r^{\theta} \rangle=\frac {1}{T} \sum_{\theta=\theta_{l}}^{\theta=\theta_{T}} \bar r^{\theta}$ for $T$ different entropy orientations. Finally, note that entropy may be normalized to the interval $[0,1]$ by multiplying $r^{k,\theta}$ by the normalizing constant $1/\log_{2}d$ in Eq. 3, provided that $0\le r^{k,\theta}\le \log_{2} d$.
As we will show in the next section, the measure defined by Eq. (5) and the image noise are inversely correlated. This opens up the possibility of using this measure as a quality assessment index for denoising algorithms. 

\subsection{Anisotropy and image noise content. Experimental results.}
Let consider a test image as the one shown in Fig. 2. This image is proposed here as the good quality visual representation of the scene and corresponds to a digital gray image coded with 8 bits, 512 by 512 pixels in size.  By corrupting this image with an increasing amount of Gaussian noise we can obtain a set of noisy images whose peak signal to noise ratio (PSNR) as presented in Table 1.
The PSNR is a common logarithmic measure of quality for images. It may be defined through the mean square error (MSE). Given two  gray-level images X and Y where the former represents the reference and the later is a noisy approximation of the first one, the MSE is defined as:
\begin{equation}
MSE=\frac{1}{M{\times}N}\sum_{i=1}^{M}\sum_{j=1}^{N}[X(i,j)-Y(i,j)]^2
\end{equation}
and from here, the PSNR is defined as
\begin{equation}
PSNR=10\log_{10}(\frac{MAX^2}{MSE})
\end{equation}
where $MAX$  is the maximum possible value in the test images, i.e.: 255 for 8 bits images. As usual, we will express PSNR in decibels (dB).

\begin{figure}[h!]
  \centering
      \includegraphics[width=0.5\textwidth]{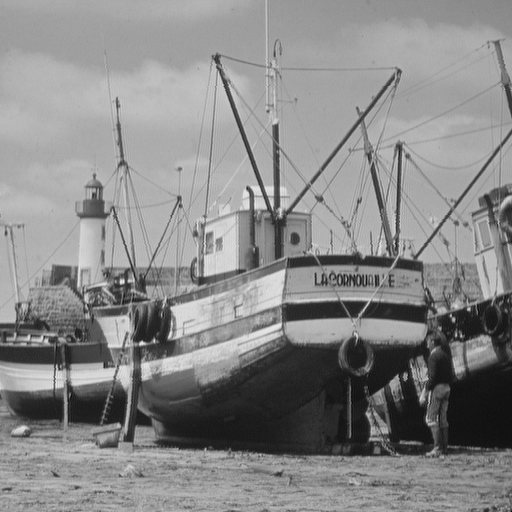}
  \caption{Test original image ('Boat')}
\end{figure}

Applying the stack filtering approach previously described in section 2, using an analysis window of $9{\times}9$ pixels for $D_{\theta}$   in Eq. (2), the six orientations indicated in Fig. 1, and $\alpha=3$  in Eq. (3), the anisotropy of each stack binary level has been measured, attaining the values represented in Fig. 3.

\begin{table}[htdp]
\caption{Gaussian noise test images }
\begin{center}
\begin{tabular}{|c|c|c|c|c|c|c|}
\hline
Image & Boat & 1 & 2 & 3 & 4 & 5 \\
\hline
PSNR & $\infty$ & 17,25 & 14,59 & 13,18 & 12,28 & 11,66 \\
\hline
\end{tabular}
\end{center}
\label{gauss}
\end{table}%

\begin{figure}[h!]
  \centering
      \includegraphics[width=0.7\textwidth]{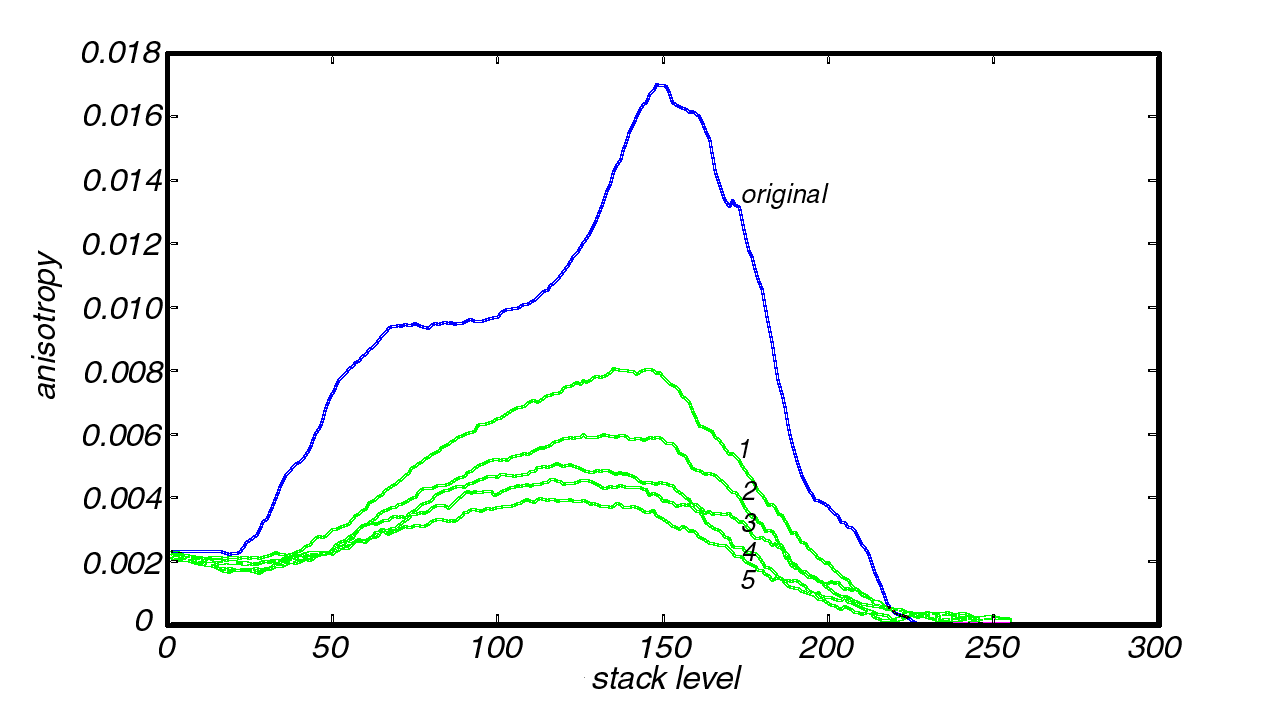}
  \caption{Anisotropy values for the different stack filter levels and different Gaussian noise content of boat image. In each level the maximum anisotropy corresponds to the original image.}
\end{figure}

A second example may be derived from the boat image but corrupting it now with different degrees of speckle noise. This is a multiplicative variety of noise affecting, for example, Positron Emission Tomography (PET) and Synthetic Aperture Radar (SAR) images. The relative quality of the test noisy images is indicated by the PSNR values presented in Table 2. Fig. 4 shows the resulting stack anisotropy by means of a graphical representation of the evolution in each binary stack level. Again the anisotropy of the original image exceeds the values given by the rest of the degraded images.

\begin{table}[htdp]
\caption{Speckle noise test images }
\begin{center}
\begin{tabular}{|c|c|c|c|c|c|c|}
\hline
Image & Boat & 1 & 2 & 3 & 4 & 5 \\
\hline
PSNR & $\infty$ & 21.86 & 18.91 & 17.40 & 16.46 & 15.79 \\
\hline
\end{tabular}
\end{center}
\label{gauss}
\end{table}%

\begin{figure}[h!]
  \centering
      \includegraphics[width=0.7\textwidth]{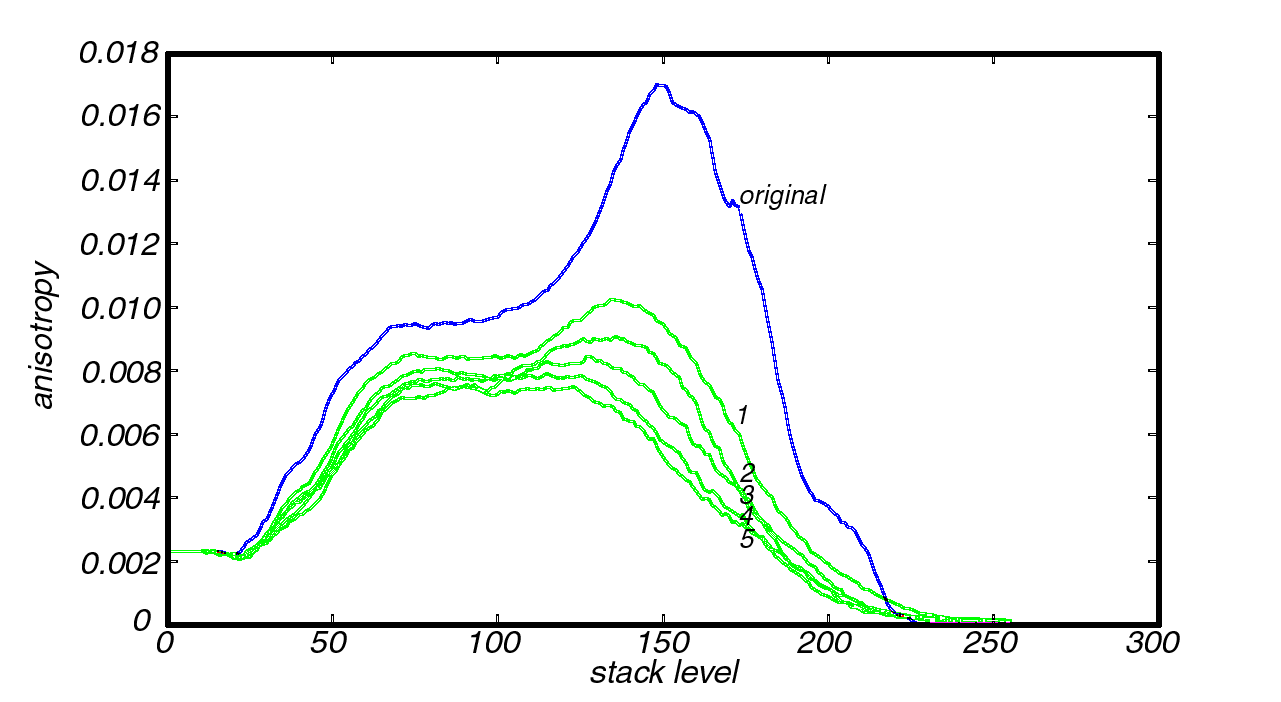}
  \caption{Anisotropy values for the different stack filter levels and different speckle noise content of boat image. In each level the maximum anisotropy corresponds to the original image.}
\end{figure}

This experiment was repeated with a third kind of noise, i.e.: impulsive noise, with similar results. Table 3 and Fig. 5 summarize such results. In this case the differences from the original image to the noisy versions are smaller. This justifies the fact that impulsive noise can be more easily removed from images than Gaussian or speckle noise, e.g.: by median filters. The method has been statistically tested with the 36 real world digital images database used in [1] in order to determine its generality.  The method has been applied following the same scheme used with the previous individual examples. Average results are shown in Table 4 and Fig. 6 for the speckle noise case, but similar results have been observed with Gaussian and impulsive noise degradations.

\begin{table}[htdp]
\caption{Impulsive noise test images }
\begin{center}
\begin{tabular}{|c|c|c|c|c|c|c|}
\hline
Image & Boat & 1 & 2 & 3 & 4 & 5 \\
\hline
PSNR & $\infty$ & 19.27 & 16.41 & 14.73 & 13.57 & 12.68 \\
\hline
\end{tabular}
\end{center}
\label{gauss}
\end{table}%

\begin{figure}[h!]
  \centering
      \includegraphics[width=0.7\textwidth]{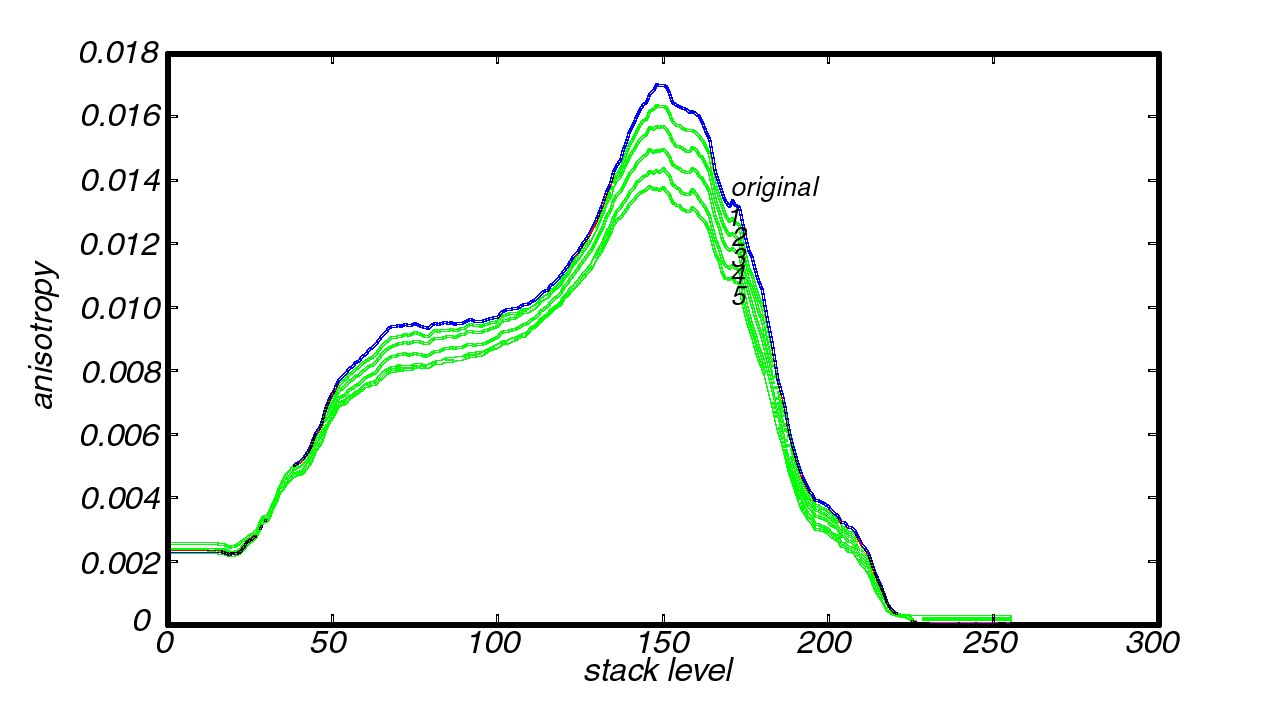}
  \caption{Anisotropy values for the different stack filter levels and different impulsive noise content of boat image. In each level the maximum anisotropy corresponds to the original image.}
\end{figure}

\begin{table}[htdp]
\caption{Set of 36 speckle noise test images (average)}
\begin{center}
\begin{tabular}{|c|c|c|c|c|c|c|}
\hline
Image & Database & 1 & 2 & 3 & 4 & 5 \\
\hline
PSNR & $\infty$ & 22.43 & 19.78 & 18.28 & 17.24 & 16.44 \\
\hline
\end{tabular}
\end{center}
\label{gauss}
\end{table}%

\begin{figure}[h!]
  \centering
      \includegraphics[width=0.7\textwidth]{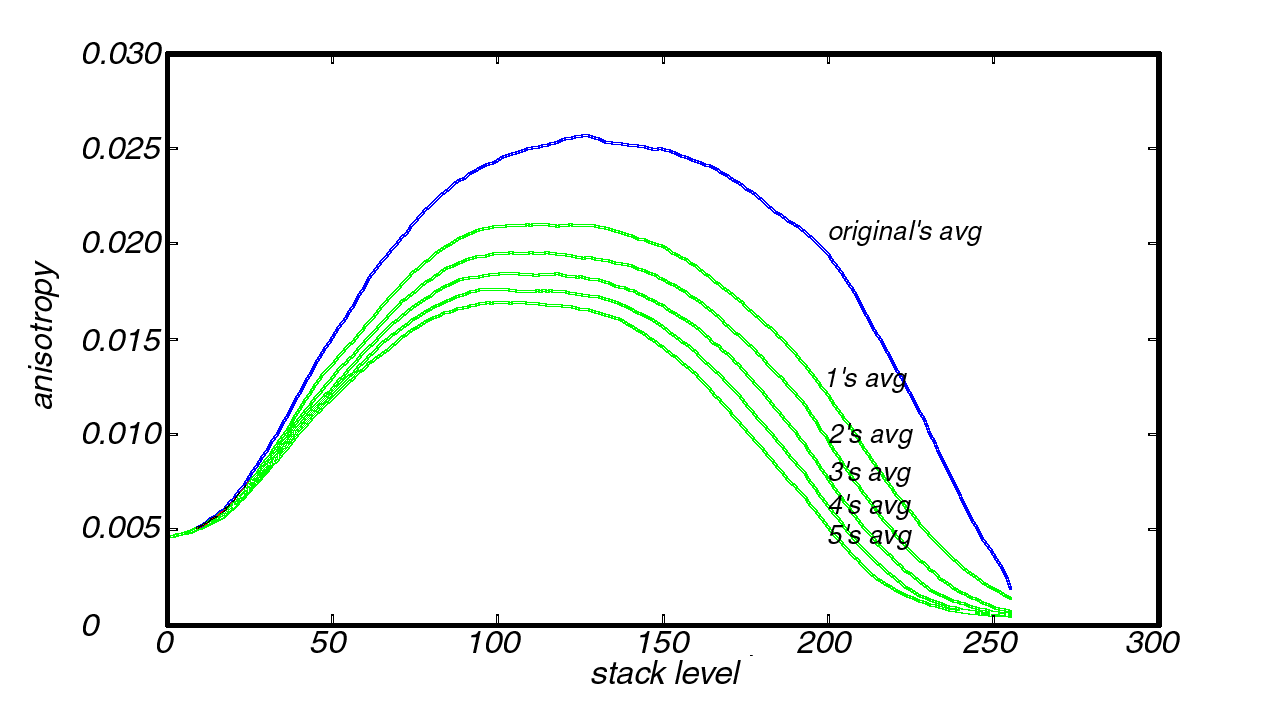}
  \caption{Averaged anisotropy values for the different stack filter binary levels with different speckle noise content of a database of 36 real-world digital images. In each level the maximum anisotropy corresponds to the averaged anisotropy of original images. The test shows how anisotropy monotonously decreases with increasing speckle noise as a general rule.}
\end{figure}

This anisotropic measure can be globalized for a gray-value image by considering the area under the graphical representation of the anisotropy.  If in each level $l$ of the stack the anisotropy $A^l$  is defined as in Eq. (5), then   
\begin{equation}
A_{G}=\sum_{l=1}^{255}A^l
\end{equation}
will measure the anisotropy for the whole image. The practical rule that may be derived is the following: {\it given a set of registered images of the same scene, the one showing the maximum value for $A_{G}$ is the preferred noise-free representation.} The quality index we propose here, based on this rule, is able to classify images within the same context in terms of their degree of noisiness. The higher the index, the lesser the presence of noise in the image. This quality index may be expressed also as a noise estimation index, by considering the inverse value of Eq. (8), but as it was defined it provides an easily to interpret quality index.  
In general, we can expect that this behavior will be followed by any kind of isotropic noise, as our experiments have confirmed. Other kind of noise, e.g. with directional patterns, may modify the anisotropy of the image but it will not necessarily diminish its value.
We can compare different results from different denoising algorithms and determining the less degraded image according to its anisotropic value, but some constraints must be considered.  We assume that the original noise is isotropic (random) and the residual noise together with the possible distortion introduced by the denoising method are also random. 

As a real-world situation, the next example illustrates the evaluation by this technique of four different speckle noise reduction methods: Frost \citep{Frost_1982}, Kuan \citep{Kuan_1987}, relaxed median filters \citep{Ben-Hamza_1999} and SRAD \citep{Yu_2002}. The test image is the gray-scale 8-bit $800{\times}800$ pixel SAR image shown in Fig. 7. The anisotropy measures using the stack filtering approach are shown in Fig. 8. From the figure we grasp an intuitive impression of the enhancement produced by the different denoising methods. Table 5 gives the global evaluation $A_{G}$ of the different outcomes obtained by applying the anisotropic index defined in (8). The second row in Table 5 shows the percentage of enhancement due to each noise reduction method. 

\begin{figure}[h!]
  \centering
      \includegraphics[width=0.5\textwidth]{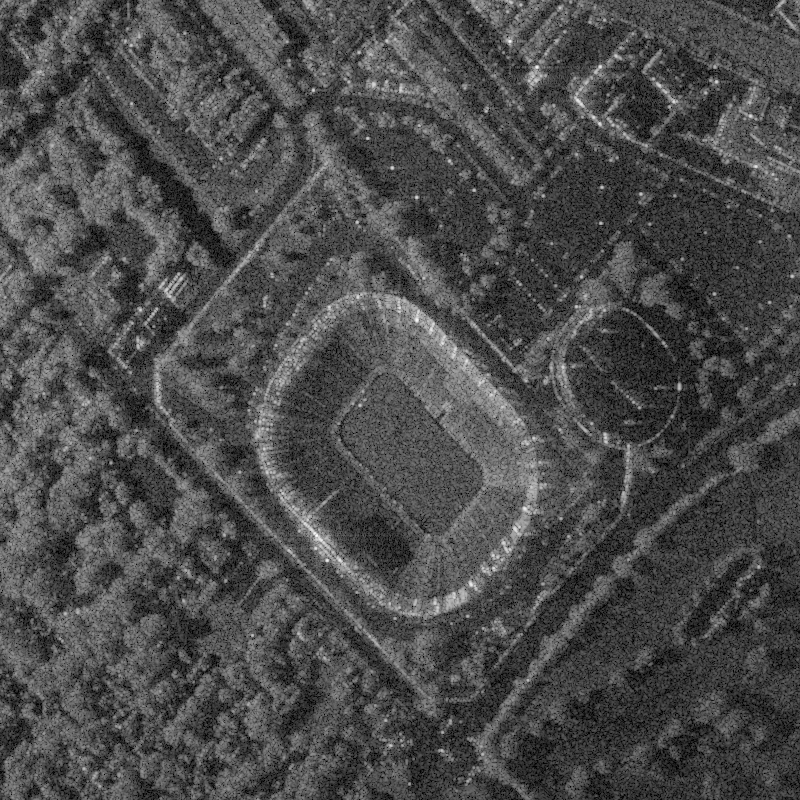}
  \caption{Example of a SAR image. This image represents a satellite SAR image degraded by speckle noise. Image courtesy from B.Brower (ITT Industries)}
\end{figure}

\begin{table}[htdp]
\caption{Real world speckle noisy image}
\begin{center}
\begin{tabular}{|c|c|c|c|c|c|}
\hline
Image & input & $R_{med}$ & Frost & SRAD & Kuan \\
\hline
                 $A_{G}$ & 0.67 & 0.71 & 0.73 & 0.76 & 0.77  \\
$\% \Delta A_{G} $& 0 & 5.97 & 8.96 & 13.43 & 14.93  \\
\hline
\end{tabular}
\end{center}
\label{gauss}
\end{table}%

\begin{figure}[h!]
  \centering
      \includegraphics[width=0.7\textwidth]{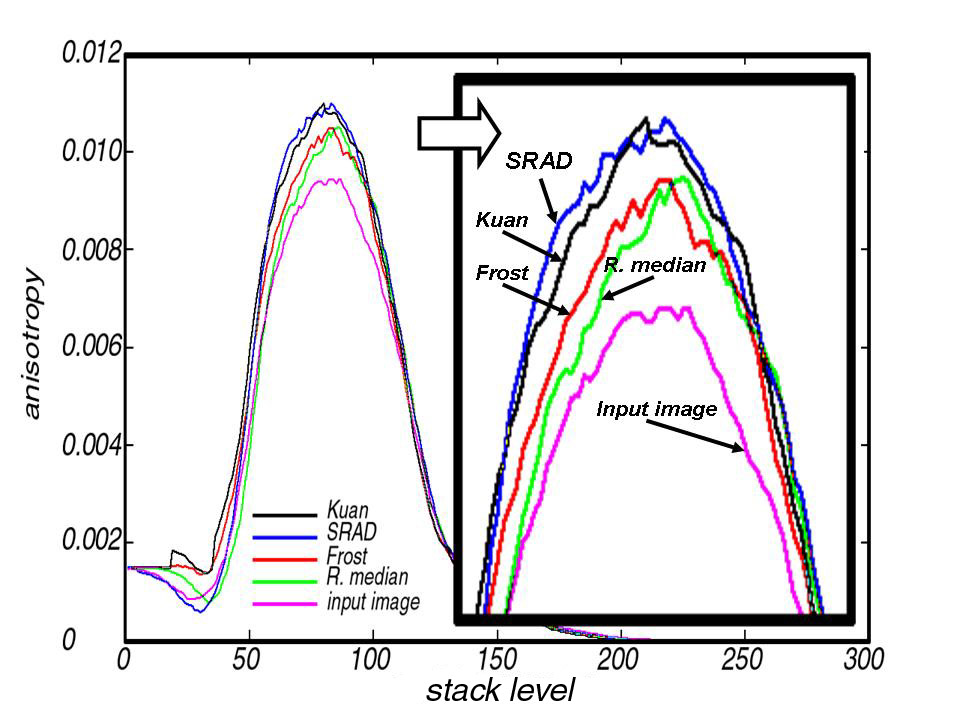}
  \caption{Averaged anisotropy values for the different stack filter binary levels for different speckle noise reduction methods applied over the test image shown in Fig. 7. The four methods are much alike but there are some differences. The areas under the curves are used as a quality index (see Eq. (8)) for each process and these values are compared in Table 5. }
\end{figure}

\section{Conclusions}
In this paper a new anisotropic no-reference measure for image denoising assessment has been introduced. Stack filters provide a binary partial representation of the information content of an image. We have presented here some experimental results indicating that anisotropy decreases with when noise increases in all binary levels of stack filters. This provides a new quality assessment tool for noise reduction algorithms. By measuring anisotropy as a function of the stack levels we have defined a quality assessment index for images. This anisotropic measure is able of handling the image information in an analytical way that may be valuable for future applications as, for example, image fusion methods. By this anisotropic measurement model, differences on quality at different image gray-levels may be determined, indicating which pixels have to be selected in the fusion process from two or more images, according to their particular anisotropic values. Further work will consider how this global anisotropic measure may be converted to a local measure for developping new filtering methods for image denoising and to study the combined influence of blur and noise. 

\section*{Acknowledgments}
This research has been partially funded by the Spanish Ministry of Science and 
Innovation from the pro jects TEC2010-20307 and TEC2010-09834-E.

\bibliographystyle{model5-names}
\bibliography{speckled}

\begin{thebibliography}{11}
\expandafter\ifx\csname natexlab\endcsname\relax\def\natexlab#1{#1}\fi
\providecommand{\bibinfo}[2]{#2}
\ifx\xfnm\relax \def\xfnm[#1]{\unskip,\space#1}\fi
\bibitem[{Astola(1997)}]{Astola_97}
\bibinfo{author}{Astola, J.} (\bibinfo{year}{1997}).
\newblock {\it \bibinfo{title}{Fundamentals of Nonlinear Digital Filtering}\/}.
\newblock \bibinfo{address}{Boca Raton, FL}: \bibinfo{publisher}{CRC Press}.
\bibitem[{Ben~Hamza et~al.(1999)Ben~Hamza, Luque-Escamilla, Mart\'{i}nez-Aroza
  \& Rom\'an-Rold\'an}]{Ben-Hamza_1999}
\bibinfo{author}{Ben~Hamza, A.}, \bibinfo{author}{Luque-Escamilla, P.},
  \bibinfo{author}{Mart\'{i}nez-Aroza, J.}, \&
  \bibinfo{author}{Rom\'an-Rold\'an, R.} (\bibinfo{year}{1999}).
\newblock \bibinfo{title}{Removing noise and preserving details with relaxed
  median filters}.
\newblock {\it \bibinfo{journal}{Journal of Mathematical Imaging and
  Vision}\/},  {\it \bibinfo{volume}{11}\/}, \bibinfo{pages}{161--177}.
\bibitem[{Buemi et~al.(2010)Buemi, Jacobo \& Mejail}]{Buemi_10}
\bibinfo{author}{Buemi, M.~E.}, \bibinfo{author}{Jacobo, J.}, \&
  \bibinfo{author}{Mejail, M.} (\bibinfo{year}{2010}).
\newblock \bibinfo{title}{{SAR} image processing using adaptive stack filter}.
\newblock {\it \bibinfo{journal}{Pattern Recognition Letters}\/},  {\it
  \bibinfo{volume}{31}\/}, \bibinfo{pages}{307--314}.
\bibitem[{Coyle \& Lin(1988)}]{Coyle_88}
\bibinfo{author}{Coyle, E.}, \& \bibinfo{author}{Lin, J.-H.}
  (\bibinfo{year}{1988}).
\newblock \bibinfo{title}{Stack filters and the mean absolute error criterion}.
\newblock {\it \bibinfo{journal}{IEEE Trans. Acoust. Speech Signal
  Process.}\/},  {\it \bibinfo{volume}{36}\/}, \bibinfo{pages}{1244--1254}.
\bibitem[{Coyle et~al.(1989)Coyle, Lin \& Gabbouj}]{Coyle_89}
\bibinfo{author}{Coyle, E.}, \bibinfo{author}{Lin, J.-H.}, \&
  \bibinfo{author}{Gabbouj, M.} (\bibinfo{year}{1989}).
\newblock \bibinfo{title}{Optimal stack filtering and the estimation and
  structural approaches to image processing}.
\newblock {\it \bibinfo{journal}{IEEE Trans. Acoust. Speech Signal
  Process.}\/},  {\it \bibinfo{volume}{37}\/}, \bibinfo{pages}{2037--2066}.
\bibitem[{Frost et~al.(1982)Frost, Stiles, Shanmugan \& Holtzman}]{Frost_1982}
\bibinfo{author}{Frost, V.}, \bibinfo{author}{Stiles, J.~A.},
  \bibinfo{author}{Shanmugan, K.~S.}, \& \bibinfo{author}{Holtzman, J.~C.}
  (\bibinfo{year}{1982}).
\newblock \bibinfo{title}{A model for radar images and its application to
  adaptive digital filtering of multiplicative noise}.
\newblock {\it \bibinfo{journal}{IEEE Trans. on Pattern Analysis and Machine
  Intell.}\/},  {\it \bibinfo{volume}{4}\/}, \bibinfo{pages}{157--166}.
\bibitem[{Gabarda \& Cristobal(2007)}]{Gabarda07}
\bibinfo{author}{Gabarda, S.}, \& \bibinfo{author}{Cristobal, G.}
  (\bibinfo{year}{2007}).
\newblock \bibinfo{title}{Blind image quality assessment through anisotropy}.
\newblock {\it \bibinfo{journal}{J. Opt. Soc. Am. A}\/},  {\it
  \bibinfo{volume}{24}\/}, \bibinfo{pages}{B42--51}.
\bibitem[{Kuan et~al.(1987)Kuan, Sawchuk, Strand \& Chavel}]{Kuan_1987}
\bibinfo{author}{Kuan, D.~T.}, \bibinfo{author}{Sawchuk, A.~A.},
  \bibinfo{author}{Strand, T.~C.}, \& \bibinfo{author}{Chavel, P.}
  (\bibinfo{year}{1987}).
\newblock \bibinfo{title}{Adaptive restoration of images with speckle}.
\newblock {\it \bibinfo{journal}{IEEE Trans. on Acoust. Speech Signal
  Proc.}\/},  {\it \bibinfo{volume}{35}\/}, \bibinfo{pages}{373--383}.
\bibitem[{R\'enyi(1976)}]{Renyi76}
\bibinfo{author}{R\'enyi, A.} (\bibinfo{year}{1976}).
\newblock \bibinfo{title}{Some fundamental questions of information theory}.
\newblock In \bibinfo{editor}{P.~Tur\'an} (Ed.), {\it
  \bibinfo{booktitle}{Selected Papers of Alfr\'ed R\'enyi}\/} chapter
  \bibinfo{chapter}{(Originally: MTA III. Oszt. Kozl., 10, 1960, pp. 251-282)}.
\newblock \bibinfo{address}{Budapest}: \bibinfo{publisher}{Akad\'emiai
  Kiad\'o.}
\bibitem[{Shannon \& Weaver(1949)}]{Shannon_49}
\bibinfo{author}{Shannon, C.}, \& \bibinfo{author}{Weaver, W.}
  (\bibinfo{year}{1949}).
\newblock {\it \bibinfo{title}{The Mathematical Theory of Communication}\/}.
\newblock \bibinfo{address}{Urbana, Chicago, London}: \bibinfo{publisher}{The
  University of Illinois Press}.
\bibitem[{Yu \& Acton(2002)}]{Yu_2002}
\bibinfo{author}{Yu, Y.}, \& \bibinfo{author}{Acton, S.~T.}
  (\bibinfo{year}{2002}).
\newblock \bibinfo{title}{Speckle reducing anisotropic difusion}.
\newblock {\it \bibinfo{journal}{IEEE Trans. on Image Proc.}\/},  {\it
  \bibinfo{volume}{11}\/}, \bibinfo{pages}{1260--1270}.

\end{thebibliography}







\end{document}